\pdfoutput=1

\documentclass[11pt]{article}

\usepackage[]{acl}

\usepackage{times}
\usepackage{latexsym}
\usepackage{booktabs}
\usepackage{graphicx}
\usepackage{hyperref}
\usepackage{amsmath}
\usepackage{amsfonts}
\usepackage{amssymb}
\usepackage{bm}
\usepackage{multirow}
\usepackage{float}
\usepackage{subcaption}
\usepackage{lscape}
\usepackage{enumitem}

\usepackage[T1]{fontenc}

\usepackage[utf8]{inputenc}

\usepackage{microtype}

\newcommand\dsb{{\sc DsBias}}
\newcommand\dsn{{\sc DsNorm}}
\newcommand\dsbbf{{\sc \textbf{DsBias}}}
\newcommand\dsnbf{{\sc \textbf{DsNorm}}}

\title{Modular Domain Adaptation}

\author{
  Junshen K. Chen \\
  Stanford University \\
  \texttt{kevinehc@gmail.com} \\\And
  Dallas Card \\
  University of Michigan \\
  \texttt{dalc@umich.edu}  \\\And
  Dan Jurafsky \\
  Stanford University \\
  \texttt{jurafsky@stanford.edu} \\}

\begin{document}
\maketitle

\begin{abstract}

Off-the-shelf models are widely used by computational social science researchers to measure properties of text, such as sentiment.
However, without access to source data it is difficult to account for domain shift, which represents a threat to validity.
Here, we treat domain adaptation as a modular process that involves separate model producers and model consumers, and show how they can independently cooperate to facilitate more accurate measurements of text.
We introduce two lightweight techniques for this scenario, and demonstrate that they reliably increase out-of-domain accuracy on four multi-domain text classification datasets when used with linear and contextual embedding models.
We conclude with recommendations for model producers and consumers, and release models and replication code to accompany this paper.

\end{abstract}
\section{Introduction}
\label{sec:introduction}

Machine learning models for tasks like sentiment analysis and hate speech detection are becoming increasingly ubiquitous as off-the-shelf tools, including as commercial packages or cloud-based APIs.
Among other applications, these models are widely used by computational social scientists to obtain standardized measurements of various document properties at scale. However, the problem of domain shift represents a threat to validity, one which is difficult for practitioners to overcome, especially without access to source data---which may be unavailable for reasons of privacy, copyright, or commercial interests.
In this paper, we propose to treat domain adaptation as a \emph{modular} process involving both \emph{model producers} and \emph{model consumers}, and show how both parties can independently cooperate to produce more reliable measurements.

\begin{figure}
\centering
\includegraphics[width=0.98\linewidth]{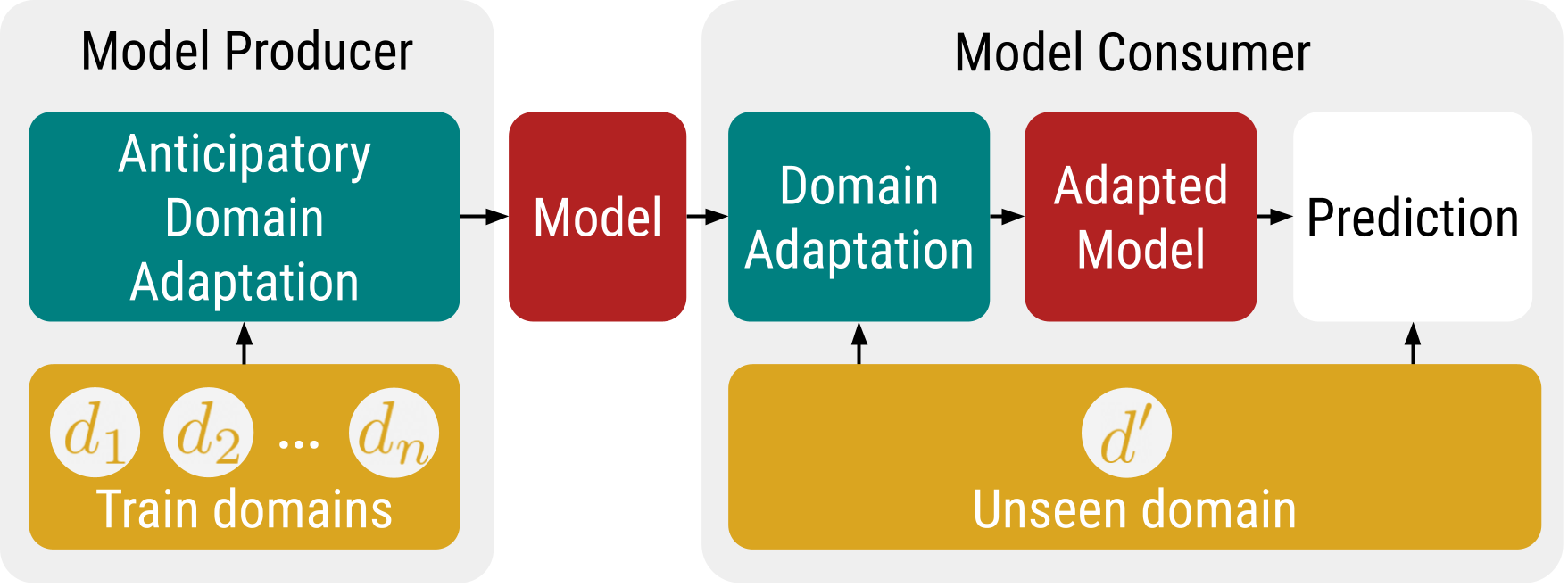}  
\caption{
Modular domain adaptation involves both model producers and model consumers, cooperating via a standardized model.
}
\label{fig:paradigm-diagram}
\end{figure}

Although this framework applies to any application involving independent model producers and consumers, we focus here on text-based instruments, including both lexicons and supervised text classification models. Using multiple datasets and baselines, we show that model consumers can obtain more accurate results by using models designed to be lightly adapted, and that model producers can facilitate such adaptation, even without providing access to source data, using what we call \emph{anticipatory domain adaptation} (see Figure \ref{fig:paradigm-diagram}). 

We introduce two techniques under this new paradigm: domain-specific bias (\dsb) and domain-specific normalization (\dsn). These methods enable model consumers to incorporate information from their domain of  interest---without additional training or hyperparameter tuning---and provide reliably better out-of-domain accuracy for both linear and contextual embedding classifiers.

In summary, this paper makes the following contributions:
\begin{itemize}
    \item We present \emph{modular domain adaptation} as a process that involves both model producers and model consumers (\S\ref{sec:problem}).
    \item We introduce two simple techniques for \textit{anticipatory domain adaptation} -- that is, ways in which model producers can facilitate adaptation by model consumers (\S\ref{sec:techniques}).
    \item We quantify the relative out-of-domain performance of linear and contextual embedding models in combination with various adaptation techniques on multiple datasets (\S\ref{sec:experiments}).
    \item We release linear and contextual models for measuring \emph{framing} in text based on the Media Frames Corpus \citep{card.2015}.\footnote{\href{https://github.com/jkvc/modular-domain-adaptation}{https://github.com/jkvc/modular-domain-adaptation}}
\end{itemize}

\section{Background and Related Work}
\label{sec:background}

There is an extensive literature on using text as data in computational social science (CSS) to study political communication, mental health, and many other  social phenomena \citep{grimmer.2013,fulgoni.2016,eichstaedt.2018,saha.2019,li.2020.covid,jaidka.2020,nguyen.2020}.
The overarching requirement in much of this work is to convert raw text (from speeches, articles, tweets, etc.) into a quantitative representation capturing some property of interest, such as sentiment or affect \citep{hatzivassiloglou.1997,huettner.2000,hutto.2014}.
Although some researchers develop bespoke models for specialized applications, those studying similar phenomena often make use of a shared set of tools, in principle allowing for comparison across studies. 

Among the most commonly used instruments are lexicons such as LIWC \citep{liwc.2010}, EmoLex \citep{mohammad.2013}, and the Moral Foundations Dictionary \citep{frimer.2019}, which offer simple, reproducible, and interpretable measurements, despite being insensitive to context.\footnote{In this paper, we use ``lexicon'' to refer to weighted or unweighted words lists corresponding to categories of interest.} Although lexicons are often developed without the use of machine learning, we can treat them interchangeably with linear models, as they are typically utilized by summing the presence of the listed features (i.e., words). The output of such models is thus a score for each document, allowing for comparisons between groups of documents, such as across time, sources, or treatment groups. Importantly, these scores should be thought of as proxies for theoretical constructs of interest, such as sentiment or ideology, to which they provide a noisy approximation \citep{jacobs.2021,pryzant.2021}.\footnote{Although lexicons are often used to obtain real-valued scores, rather than as classifiers, we assume for the sake of simplicity that any available in-domain annotations are collected as categorical labels, and evaluate all models as classifiers, using an appropriate threshold where necessary.} 

Although open source models have numerous advantages for research, model creators may be unable or unwilling to share the data that their models are based on, especially for commercial lexicons, like LIWC, and cloud-based products like Perspective API.\footnote{\href{https://www.perspectiveapi.com}{https://www.perspectiveapi.com}} 
Despite their limitations, these systems provide convenient, comparable, and easy-to-use tools for CSS researchers.
However, those who use such models face the dual problems of 1) adapting them to a new domain; and 2) assessing validity in that domain, and will often want to do so with relatively constrained resources.

Domain adaptation is an important area of research within machine learning, but most work tends to assume either access to source data (e.g., for re-weighting; \citealp{huang.2006,jiang.2007,azizzadenesheli.2019}), or extensive labeled data in the new domain. For contextual embedding models in NLP, continued training on a small amount of labeled data offers benefits \citep{radford.2017,howard.2018}, though this requires sufficient data for fine-tuning, validation, and evaluation (to assess performance in the target domain), as well as access to sufficient computational resources (typically GPUs). 

Self-training (augmenting source data using predicted labels in the new domain) provides an alternative strategy, and has been shown to work both theoretically and practically \citep{kumar.2020}, but typically assumes access to the original source data, and requires making choices about multiple hyperparameters, which is difficult in the absence of extensive validation data. A few papers have considered the problem of domain adaptation without source data \citep{chidlovskii.2016,liang.2020}, but tend to emphasize resource-intensive solutions (e.g., using GANs; \citealp{li.2020}).

A different but related paradigm is ``deconfounded lexicon induction'' \citep{pryzant-etal-2018-interpretable, pryzant.2018}, where the goal is to learn a model that accounts for the influence of non-textual attributes (such as domain). 
Because this approach tries to eliminate the influence of confounders, we might expect it to produce a more domain-agnostic model, and we therefore include experiments with the proposed techniques for the purpose of comparison.

\section{Methods}
\label{sec:methods}

\subsection{Problem Formulation}
\label{sec:problem}

In this work, we make the distinction between \emph{model producers} and \emph{model consumers}. Model producers wish to train a model on a labeled dataset of documents coming from one or more domains (e.g., political issues, or paper categories), where each document, $\mathbf{x}_i$, has an associated categorical class label, $y_i \in \mathcal{Y}$, as well as a domain, $d_i \in \mathcal{D}$.
Model consumers, by contrast, will apply the trained model to a new domain, $d' \notin \mathcal{D}$, without access to either the source data or extensive labeled data from their domain of interest.\footnote{We assume that typical model consumers in CSS are capable of generating some labeled data in their domain (e.g., by manually annotating data), but have insufficient resources available to create a large labeled dataset.}

Note that in our setup, the producer and consumer have different goals and face different constraints. 
The model producer's goal is to create a self-contained model, without sharing any source data associated with training, due to reasons such as privacy, copyright, or commercial interests.

The model consumer's goal, by contrast, is to achieve high accuracy in a new domain, $d'$, without needing extensive resources for either labeling data or training a new model. 
Especially for applications in CSS, we also assume that model consumers will need to estimate accuracy in their domain, as part of demonstrating validity \citep{jacobs.2021}.

In this paper, we compare the performance under these constraints of two especially common approaches to creating text classification models---logistic regression with bag-of-words features and contextual embedding models---and propose two methods (\dsb and \dsn; \S\ref{sec:techniques}) by which model producers can facilitate domain adaptation by model consumers.

\subsection{Underlying Models}

As foundations from which to experiment with techniques for modular domain adaptation, we make use of two standard baseline approaches in text classification: regularized logistic regression and fine-tuned contextual embedding models. In both cases, the model
is trained using an appropriate loss function (e.g., logistic or cross entropy), computed with respect to predicted probabilities:
\begin{align}
\mathbf{\hat p}_i &= \textrm{softmax}(\mathbf{b} + f(\mathbf{x}_i)^\top  \mathbf{W} ) \label{eq:phat}
\end{align}
where $\mathbf{b} \in \mathbb{R}^k$ is a bias vector, $\mathbf{W}$ is an $h \times k$ weight matrix, $f(\cdot)$ encodes a document as an $h$-dimensional vector, and $\mathbf{\hat p}_i \in \Delta^k$ is the predicted distribution over $k$ classes.\footnote{Or equivalently for binary labels: a logistic function instead of a softmax, $p_i \in [0,1]$, $b \in \mathbb{R}$, and $\mathbf{w} \in \mathbb{R}^h$.}

For logistic regression, $f(\cdot)$ encodes $\mathbf{x}_i$ as a sparse bag-of-words vector, with $h$ equal to the size of the vocabulary. 
For contextual embedding models, $f(\mathbf{x}_i) \in \mathbb{R}^h$ is the penultimate dense representation produced by feeding document $i$ into a contextual embedding model, plus additional layers in the case of a multi-layer decoder.

\subsection{Deconfounding Techniques}
\label{sec:deconfounding}

To augment the underlying models, we begin with previously proposed techniques for removing the influence of domain. 
Although mainly designed to account for explicitly modeled features of the data, and not specifically focused on domain adaptation, \citet{pryzant.2018} proposed two methods for \emph{deconfounded lexicon induction}---that is, attenuating the influence of non-textual document properties, including domain, when learning an interpretable model. 
Since these are carried out solely by model producers, we use them as baselines.

\begin{figure}
\centering
\includegraphics[width=1\linewidth]{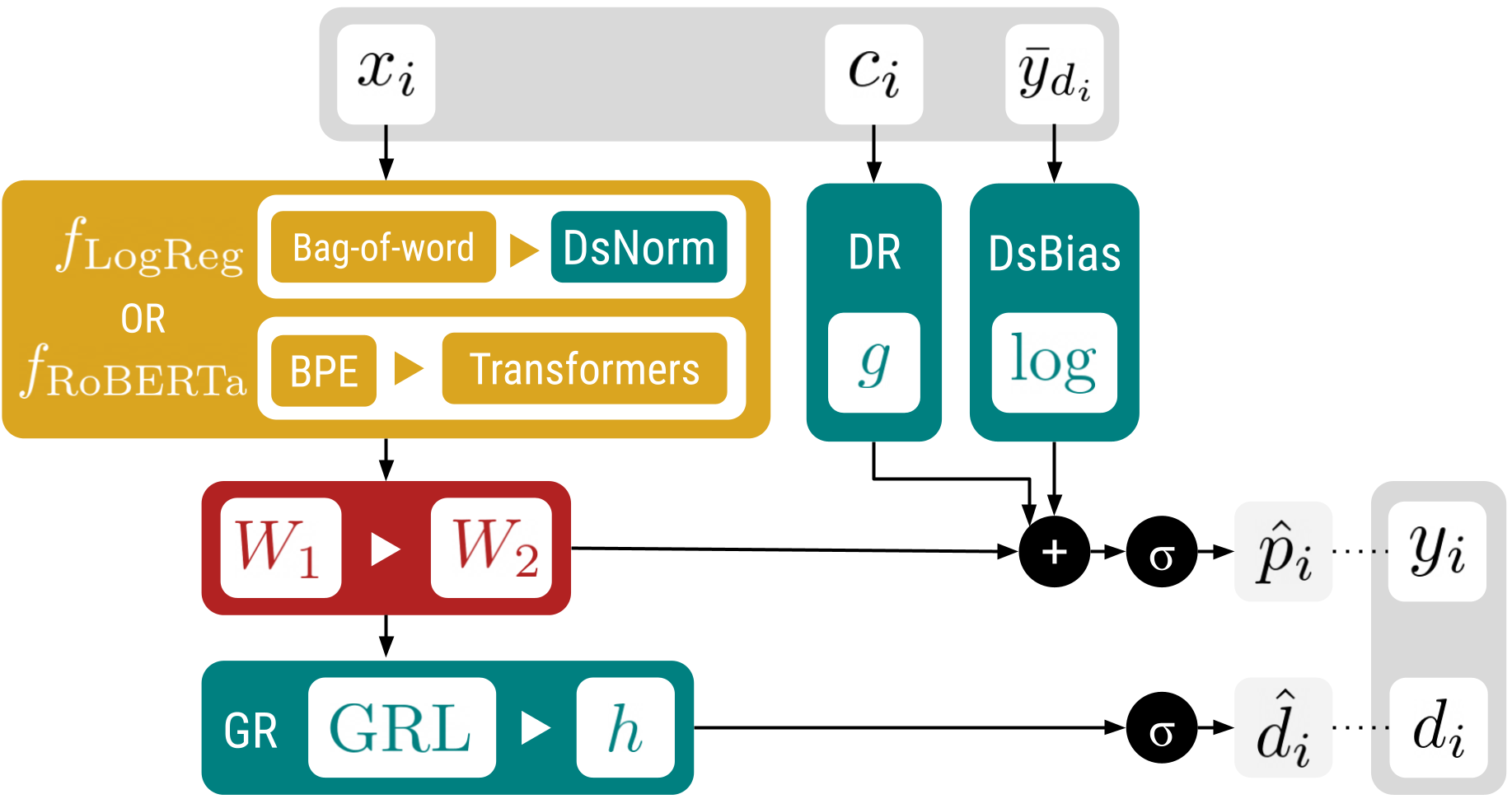}
\caption{
Model diagrams of base predictors in conjunction with proposed techniques, showing how pieces fit together. 
All deconfounding and adaptation techniques are marked in green and are optional. Base predictor is marked in yellow. 
}
\end{figure}

\paragraph{Deep Residualization (DR):} 
As one way of deconfounding labels from potential confounds, \citet{pryzant.2018} proposed learning a mapping from observable confounds to labels, and integrating that into the prediction. Specifically, we replace the bias term $\mathbf{b}$ in Eq. (\ref{eq:phat}) with an instance specific vector, i.e., 
\begin{align}
    \mathbf{\hat p}_i = \textrm{softmax}(g(\mathbf{c}_i) + f(\mathbf{x}_i)^\top \mathbf{W}),
\end{align}
where $\mathbf{c}_i$ is a vector of confounds for document $i$, and $g(\cdot)$ is a feed-forward network mapping from confounds to a dense vector representation $\in \mathbb{R}^k$. 

In our case, $\mathbf{c}_i$ is a one-hot vector representing domain (i.e., $d_i$). Since the ultimate application domain is not available at training time, the model consumer would use the domain agnostic predictor, setting $g(\mathbf{c}_i) = \mathbf{0}$ for the unseen domain.

\paragraph{Gradient Reversal (GR):}
\citet{pryzant.2018} 
also proposed using gradient reversal for deconfounding. That is, we train the model to successfully predict an instance's label, while being \emph{unable} to predict the domain. To implement this, we factorize the weight matrix $\mathbf{W}$ into two matrices, $\mathbf{W}_1$ and $\mathbf{W}_2$, and apply gradient reversal to the intermediate representation used to predict domain, i.e.
\begin{align}
 \mathbf{\hat p}_i &=  \textrm{softmax}(\mathbf{b} + (f(\mathbf{x}_i)^\top \mathbf{W}_1)^\top \mathbf{W}_2)  \\
 \mathbf{\hat d}_i &= \textrm{softmax}(h(\textrm{GRL}(f(\mathbf{x}_i)^\top \mathbf{W}_1))),
\end{align}
where $\mathbf{\hat d}_i \in \Delta^{|\mathcal{D}|}$ is the predicted distribution over domains,  $h(\cdot)$ is a feed-forward network, and GRL reverses the gradients with respect to $\mathbf{W}_1$ during training \cite{ganin-adversarial}.

\subsection{Anticipatory Adaptation Techniques}
\label{sec:techniques}

As mentioned, the above techniques were designed for deconfounding by the model producer, and not for domain adaptation by the model consumer. Here we introduce two new methods by which a model producer might facilitate adaptation, without having to share training data or requiring knowledge of the model consumer's domain.

\paragraph{Domain-Specific Bias (\dsbbf):}

A key limitation of deep residualization (DR) is that it has no way to incorporate information about a previously unseen domain. As an alternative, we modify the idea of DR by expressing the instance-specific bias in terms of the distribution of labels in the corresponding domain. This allows model consumers to inject information about a new domain into the model at prediction time, given knowledge about the relevant label distribution. 
Specifically, for each domain $d$ we set the bias term in Eq. (\ref{eq:phat}) to be the element-wise log of a vector of label frequencies in that domain, i.e., 
\begin{align}
    \mathbf{\hat p}_i &= \textrm{softmax}(\log(\mathbf{\bar y}_{d_i}) + f(\mathbf{x}_i)^\top \mathbf{W})
\end{align}
where $\mathbf{\bar y}_{d_i} \in \Delta^k$ is a vector of estimated label frequencies in the domain of instance $i$. Using the log of the estimated label frequencies means that the learned weights ($\mathbf{W}$) represent additive deviations (in log space) from baseline frequencies, much like in SAGE \citep{eisenstein.2011}.

At training time, $\mathbf{\bar y}_{d_i}$ can be estimated by the model producer from labeled data in each domain. 
At prediction time, model consumers can provide an approximate label distribution for a new domain by either estimating it from a small amount of labeled data, or by leveraging prior knowledge of the domain itself. 
Thus, \dsb~benefits from having some labeled data in the new domain, but does not require additional training by model consumers.

\paragraph{Domain-Specific Normalization (\dsnbf):}

As an additional option for linear models, and inspired by normalization techniques used in deep learning, we also consider normalizing each element in the bag-of-words feature vector according to its expected frequency of the individual domain: 
\begin{align}
    f'(\mathbf{x}_i) = f(\mathbf{x}_i) - 
    \Sigma_{j=1}^{N_{d_i}} f(\mathbf{x}_j) / N_{d_i},
\end{align}
where $f(\mathbf{x}_i)$ is a vector of feature values, and $N_{d_i}$ is the number of instances in the domain of instance $i$.
This allows for a commonly occurring word (e.g., the word ``climate'' in climate change news) to become less important if it occurs in the current domain, and relatively more important in others.\footnote{Like TF-IDF, \dsn~scales feature values based on frequency, but keeps all (binarized) feature values between $-1$ and $1$, even for rare words.}
Because this does not require labeled data, it can be applied directly to a new domain by model consumers. 

\subsection{Domain Fine-Tuning (DFT)}
\label{sec:dft}

Past work on pretrained contextual embedding models has demonstrated that continued training on labeled samples from a new domain can effectively adapt the model to that domain, improving performance \citep{radford.2017,howard.2018,gururangan.2020}.

Although powerful, there are several reasons why this may not be an option for model consumers. First, many APIs and commercial systems will not provide this functionality or expose the necessary parts of the model.
Second, the computational resources required for fine-tuning (i.e., GPUs) may be prohibitive for some users. Third, fine tuning means that individual model consumers will no longer be applying the same standardized model, thus reducing the comparability of results. Nevertheless, we include experiments with DFT in order to quantify how much better a model consumer could do with sufficient labeled data for training and evaluation in their domain (\S\ref{sec:experiments}), and compare fine tuning an off-the-shelf model to one that has been fine-tuned for the same task on out-of-domain data (\S\ref{sec:adapt}). 

\section{Experiments}
\label{sec:experiments}

In this section we systematically evaluate the performance of both underlying models in conjunction with all available techniques in section \S\ref{sec:methods}, to quantitatively evaluate their performance, and to derive best practices as advice to practitioners when applying them to real data under various settings. For simplicity, we use accuracy as the primary metric of evaluation in all our experiments.

\subsection{Data}
\label{sec:data}

Because our primary interest is to evaluate modular domain adaptation techniques, we choose datasets with instances from multiple known domains, so that we can hold out each domain in turn to estimate performance when adapting to a previously unseen domain.
In particular, we make use of four datasets in our experiments (see Table \ref{tab:datasets}): the Media Frames Corpus ({\sc mfc};  \citealp{card.2015})  the arXiv Dataset ({\sc arxiv}; \citealp{clement2019arxiv}), the Amazon Reviews Dataset ({\sc amazon}; \citealp{ni-etal-2019-justifying}), and a collection of sentiment classification datasets ({\sc senti}; see below).

{\sc \textbf{mfc}} is a dataset of news articles on 6 different issues (e.g., ``climate change''), and each article is labeled to have 1 of 15 possible primary ``frames'', which are assumed to generalize across issues. As intuition would suggest, different frames are emphasized in coverage of different issues (e.g., climate change is discussed more in terms of  ``capacity and resources'' than ``crime and punishment'').

{\sc \textbf{arxiv}} is the dataset of all scholarly articles published on \url{arXiv.org}. We consider articles in 6 categories in the taxonomy relevant to machine learning (e.g., cs.CL, ``Computation and Language''). 
For each article, we consider the year in which it was published, discretised into 4 time periods, and try to predict the time period from the abstract, using taxonomic categories as domains.\footnote{Divided by the years 2008, 2014, and 2019, which are rough markers of major machine learning milestones.}

{\sc \textbf{amazon}} is a subsampled dataset of product reviews from Amazon for the most popular 5 categories. 
Each review is associated with a review score (negative: 1; neutral: 2-4; positive: 5) which we try to predict from the review text. 

{\sc \textbf{senti}} is a collection of diverse, subsampled sentiment classification datasets: Twitter US Airline Sentiment \citep{twitter.airline}, Amazon Book Reviews \citep{ni-etal-2019-justifying}, IMDb Movie Reviews \citep{maas-EtAl:2011:ACL-HLT2011}, tweets from Sentiment 140 \citep{go2009twitter}, and the Stanford Sentiment Treebank (SST; \citealp{socher-etal-2013-recursive}). The domains included in this dataset differ from each other in various ways (e.g., IMDb reviews are often a few paragraphs long, whereas SST utterances are much shorter), which is intended to mimic scenarios in which model consumers might apply off-the-shelf sentiment analysis tools. From each sample we classify instances as positive or negative.

\begin{table}[]
    \centering
    \small
    \begin{tabular}{c c c c c}
        \toprule
        Dataset & $|\mathcal{Y}|$ & Domains & Min $N_d$ & Max $N_d$ \\
        \midrule
        \sc{mfc} & 15 & 6 & 4220 & 8898 \\
        \sc{arxiv} & 4 & 6 & 5338 & 59612 \\
        \sc{amazon} & 3 & 5 & 4199 & 22573 \\
        \sc{senti} & 2 & 5 & 3088 & 10003 \\
        \bottomrule
    \end{tabular}
    \caption{Dataset statistics, showing the number of categories (labels), domains, and minimum and maximum number of labeled instances per domain. For details of data splits, see appendix \ref{appendix:data-split}.}
    \label{tab:datasets}
\end{table}

\subsection{Implementation Details}

As a linear baseline, we use L1-regularized logistic regression (LogReg) operating on binarized bag of word features, which has been shown to be a competitive choice among similar models \citep{wang.2012}. We limit ourselves to a vocabulary of the 5000 most frequent lowercased words in the training set.
We use full-batch gradient descent to optimize the models, with L1 regularization on the weight matrices only. Regularization strength is determined for each configuration using grid search on in-domain cross validation splits, then applied to the full in-domain training set.

For contextual embedding classifiers, we use RoBERTa, fine-tuning the publicly available \texttt{roberta-base} from Hugging Face \citep{wolf2020huggingfaces}, using AdamW \citep{loshchilov2019decoupled} with a fixed dropout rate of 0.2. We use early stopping with number of epochs determined for each configuration using in-domain cross validation splits, then applied to the full in-domain training set. For additional details, please refer to Appendix \ref{appendix:hyperparams}.

\subsection{Out-of-domain Performance}
\label{sec:holdout-source}

\begin{table*}
\centering
\small
\begin{tabular}{llcccccccc}
\toprule
&& \multicolumn{2}{c}{\sc mfc} & \multicolumn{2}{c}{\sc arxiv} & \multicolumn{2}{c}{\sc amazon} & \multicolumn{2}{c}{\sc senti} 
\\
&& acc & $\sigma_\Delta$ & acc & $\sigma_\Delta$ & acc & $\sigma_\Delta$ & acc & $\sigma_\Delta$         
\\ \midrule
& Most common       & 0.276 & - & 0.526 & - & 0.631 & - & 0.495 & - 
\\ \midrule
\multirow{7}{*}{ \rotatebox[origin=c]{90}{LogReg}} 
& Base              & 0.508 & - & 0.543 & - & 0.672 & - & 0.647 & -\\
& DR                & 0.503 & 0.009 & 0.551 & 0.005 & 0.674 & 0.004 & 0.648 & 0.003 \\
& GR                & 0.500 & 0.004 & 0.541 & 0.005 & 0.709 & 0.001 & 0.638 & 0.003 \\
& \dsb~(250)         & 0.515 & 0.020 & 0.564 & 0.024 & 0.714 & 0.004 & 0.690 & 0.052  \\
& \dsn+\dsb~(250)     & 0.532 & 0.018 & 0.568 & 0.013 & 0.716 & 0.006 & 0.700 & 0.041 \\ \cmidrule{2-10}
& \dsb~(oracle)      & 0.524 & 0.022 & 0.563 & 0.013 & 0.715 & 0.003 & 0.695 & 0.041 \\
& \dsn+\dsb~(oracle)  & 0.541 & 0.015 & 0.568 & 0.012 & 0.717 & 0.002 & 0.709 & 0.039 \\ 
\midrule
\multirow{6}{*}{ \rotatebox[origin=c]{90}{RoBERTa} } 
& Base              & 0.599 & - & 0.584 & - & 0.772 & - & 0.789 & -  \\
& DR                & 0.594 & 0.014 & 0.593 & 0.007 & 0.782 & 0.017 & 0.817 & 0.012   \\
& GR                & 0.202 & 0.039 & 0.512 & 0.003 & 0.777 & 0.012 & 0.684 & 0.068  \\
& \dsb~(250)         & 0.613 & 0.030 & 0.599 & 0.010 & 0.772 & 0.036 & 0.819 & 0.016 \\
& DFT (250)         & 0.683 & 0.032 & 0.615 & 0.012 & 0.785 & 0.025 & 0.831 & 0.018 \\ \cmidrule{2-10}
& \dsb~(oracle)      & 0.622 & 0.026 & 0.600 & 0.013 & 0.779 & 0.012 & 0.819 & 0.014   \\
\bottomrule
\end{tabular}
\caption{Average out-of-domain accuracy on four datasets show consistent findings for both LogReg and RoBERTa: (1) \dsb~with the oracle label distribution offers a small but reliable gain in accuracy over the Base models; (2) gains are almost as large when approximating the oracle distribution with 250 labeled examples; (3) \dsn~also offers a small but reliable benefit for linear models when used in combination with \dsb; (4) Deconfounding techniques (DR and GR) do not improve out-of-domain accuracy over Base; (5) RoBERTa achieves much better out-of-domain accuracy than LogReg, even without fine tuning to the target domain; (6) Additional fine tuning to 250 labeled example (DFT) offers additional gains, though this may not be an option for some model consumers. $\sigma_\Delta$ is the standard deviation (across held-out domains) of the improvement over the baseline (Base).
}
\label{tab:holdout-source-acc}
\end{table*}

As our primary evaluation, we assess each technique in combination with each of our base models (LogReg vs. RoBERTa). For each domain of each dataset, we create a dedicated held-out test set. During training, for each dataset, we hold out each domain in turn, and use the remaining domains as in-domain training data.

We report average performance on  out-of-domain test sets, along with variance (across domains) in improvement over the baseline model in Table \ref{tab:holdout-source-acc}. For \dsb, we evaluate performance both when assuming oracle knowledge of the label distribution in the held-out domain, and when we estimate it from a random sample of 250 instances, which we also use for DFT.

There are four important takeaways from these results. First, RoBERTa offers a dramatic improvement over base logistic regression in out-of-domain performance (4--18\% improvement), even without additional fine-tuning by the model consumer.\footnote{As expected, both LogReg and RoBERTa show large drops in performance from the domains in which they were trained (3-10\% on average, depending on dataset; see Table  \ref{tab:perf-drop} in Appendix \ref{appendix:perf-drop}).} Thus, although some model consumers may still prefer linear models or lexicons for greater interpretability (see Appendix \ref{appendix:lexicon}), the CSS community would greatly benefit from having model producers release both linear \emph{and} contextual embedding models. Moreover, fine-tuning RoBERTa to even a small amount of in-domain labeled data produces another additional improvements (though with caveats, as discussed in \S\ref{sec:dft}).

Second, the deconfounding techniques (DR and GR) offer little or no benefit over the baseline in terms of out-of-domain performance. Thus, while they may work for removing the influence of domain in constructing a lexicon, they do not appear to produce a domain agnostic lexicon in a way that is beneficial for model consumers. 

Third, \dsb~(using the log label distribution for each domain) offers a small but reliable benefit (2-4\%) to model consumers when working with a known label distribution, and this applies to both linear and contextual embedding models. 
Moreover, this still holds when model consumers estimate this distribution from a small amount of labeled data (here 250 instances). A key advantage to \dsb~is that it requires no additional training by model consumers, and essentially keeps the underlying model unchanged, preserving comparability across studies. Moreover, estimating a low-dimensional label distribution requires relatively few samples, with statistically bounded errors given a random sample (see \S\ref{sec:estimating} below).

Fourth, \dsn~(normalizing features by domain) offers a small additional benefit when used in combination with \dsb~for linear models, and it can be applied by model consumers based purely on unlabeled data from their domain.

Based on what evaluations can be justified using a simple power analysis \citep{card.2020}, we verify that LogReg+\dsb+\dsn~is significantly better than LogReg for all but one dataset (using McNemar's test), as is RoBERTa+\dsb~compared to RoBERTa (for all datasets; see Appendix \ref{appendix:power-analysis}). Finally, in Appendix \ref{appendix:extreme-data-scarcity}, we verify that our findings hold even if the model producer is only able to train on a single domain.

\subsection{Estimating the Label Distribution}
\label{sec:estimating}

\dsb~achieved the best performance when given the oracle label distribution of the target domain, but in practice this is unlikely to be known precisely. To study the effect of using an estimated label distribution with the technique, we here assume that we only have very few labeled samples from the unseen domain. Specifically, we run the same experiment in \S\ref{sec:holdout-source} where we vary the number of samples used to estimate the label distribution in the target domain. 

\begin{figure}
    \centering
    \includegraphics[width=0.48\textwidth]{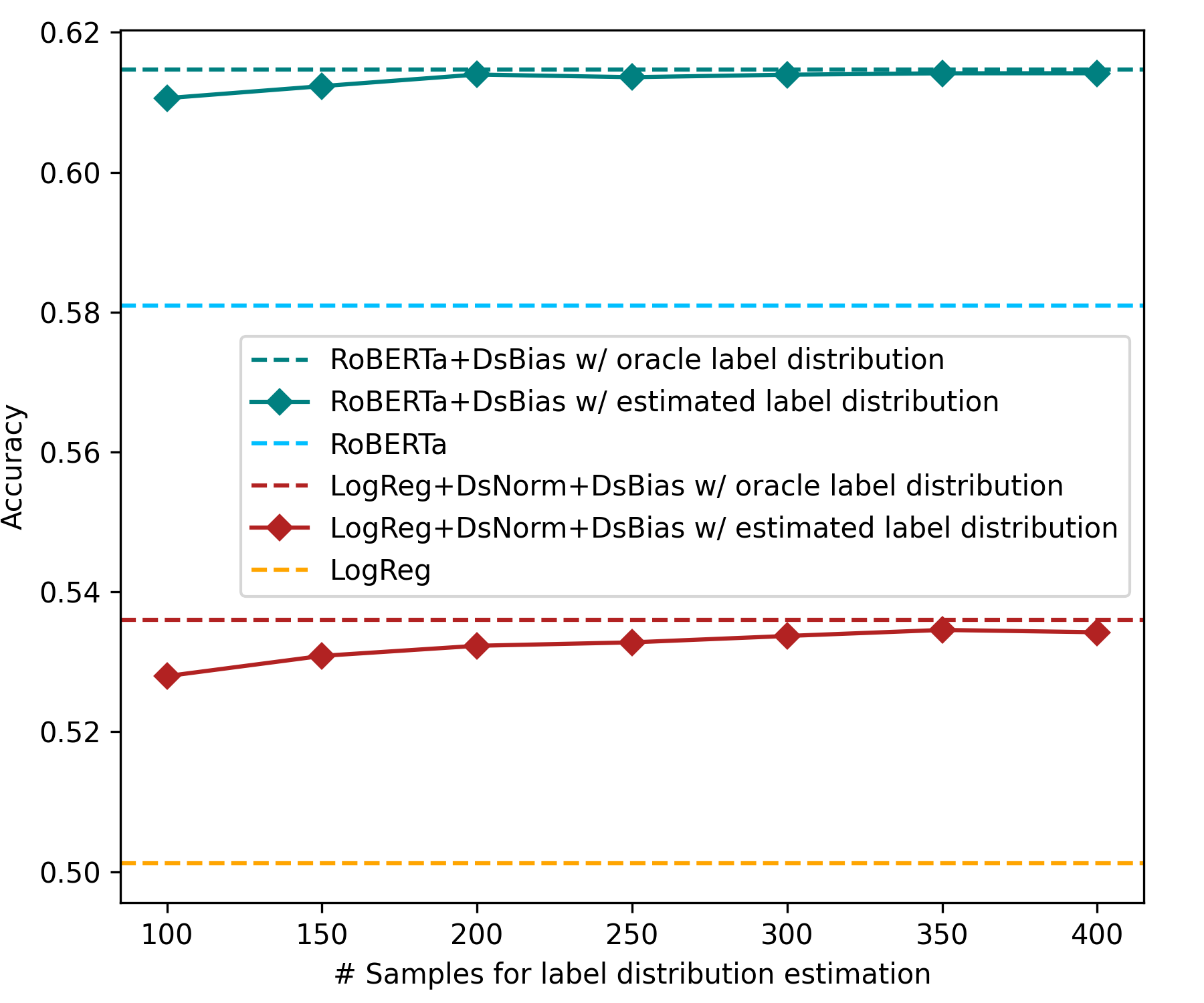}
    \caption{Average validation accuracy in unseen domains of {\sc mfc}, using a varying number of target domain samples to estimate label distribution for \dsb.
    }
    \label{fig:labelprop-est}
\end{figure}

Figure \ref{fig:labelprop-est} demonstrates that with only as few as 100 labeled samples, average performance using \dsb~improves from the base model, and arrives within 1 percent of accuracy from using the ground truth distribution. 
For each heldout domain, we run 5 trials each estimating label distribution using  a fixed number of random samples, evaluate performance on the full train set of the heldout domain, then average across all trials and all heldout domains. 
Further including more labeled samples in estimating label distribution results in marginal, upper-bounded improvements. 

\begin{figure}
    \centering
    \includegraphics[width=0.48\textwidth]{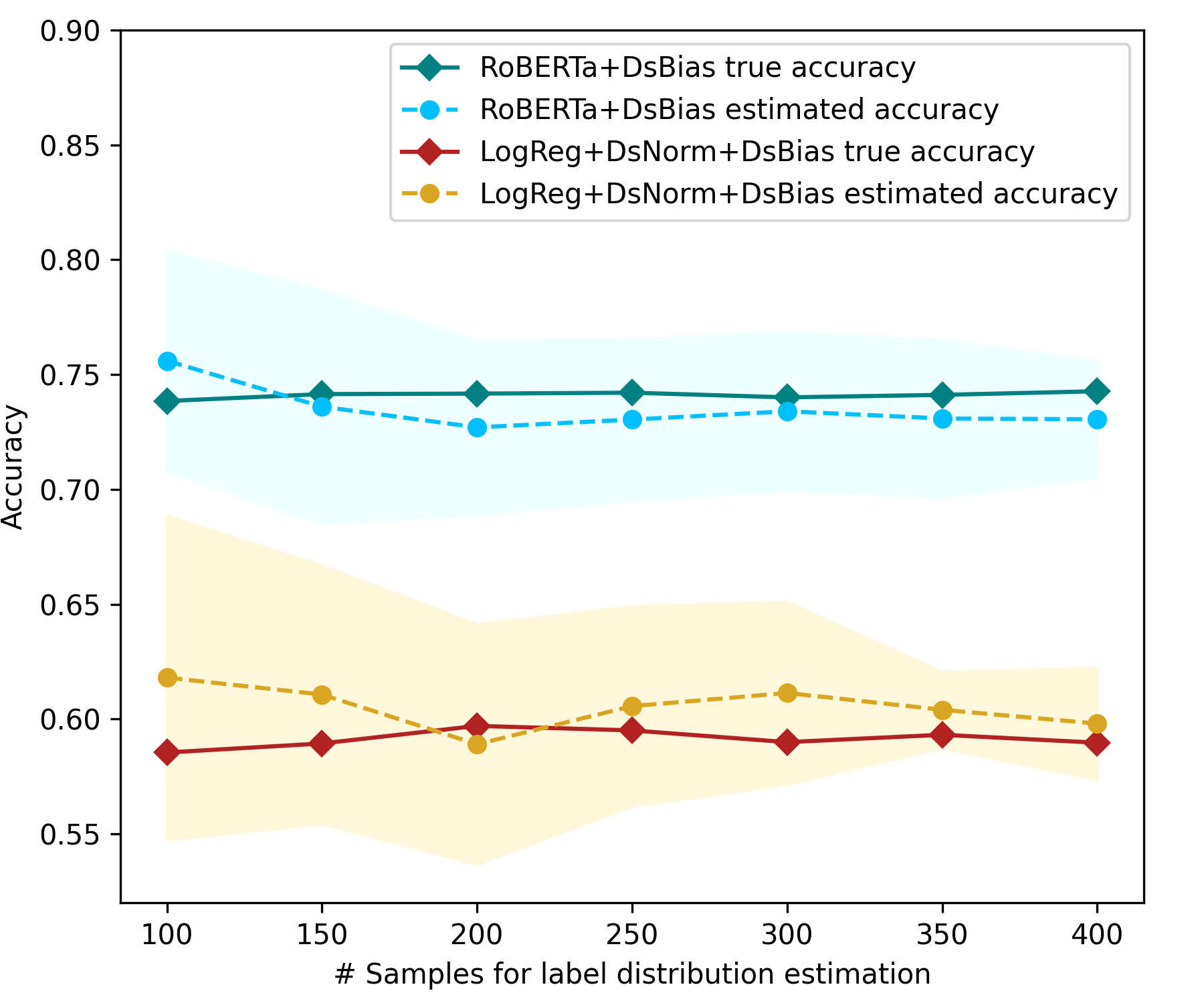}
    \caption{Validation accuracy of calculated from all holdout samples, and from limited samples, of the Sentiment 140 dataset in {\sc senti}. Shaded area denotes 1 standard deviation from mean estimated performance. For all domains in all datasets, see appendix \ref{appendix:estimate-performance}. }
    \label{fig:performance-est}
\end{figure}

Especially for CSS applications, model consumers are likely to care as much about estimating performance in their domain (to ensure validity) as they do about improving performance.
An additional advantage of \dsb~is that one can easily use two-fold estimation to effectively re-use any available labeled data for both estimating the label distribution and evaluating performance. That is, split the available labeled data in two, use half to estimate the label distribution, and the other half to estimate performance. Repeat this (reversing roles), and then take the average performance as an estimate of in-domain accuracy, without any model training or hyperparameter tuning required. One can then use all of the labeled data to estimate the label distribution for making predictions on the full unlabeled dataset. As shown in Figure \ref{fig:performance-est}, this produces an unbiased estimate, with variance that decreases with the amount of labeled data.

\subsection{Domain Fine-tuning}
\label{sec:adapt}

One major advantage of contextual embedding models like RoBERTa is that one can easily fine-tune to a new domain by simply continuing to train on additional labeled data \citep{gururangan.2020}. Although this may not be a possibility for some model consumers (see \S\ref{sec:dft}), we evaluate this approach for the sake of completion.\footnote{Importantly, contextual embedding models can easily be \emph{applied} with minimal computational requirements, but domain fine-tuning requires more resources and expertise.}

\begin{figure}
    \centering
    \includegraphics[width=0.49\textwidth]{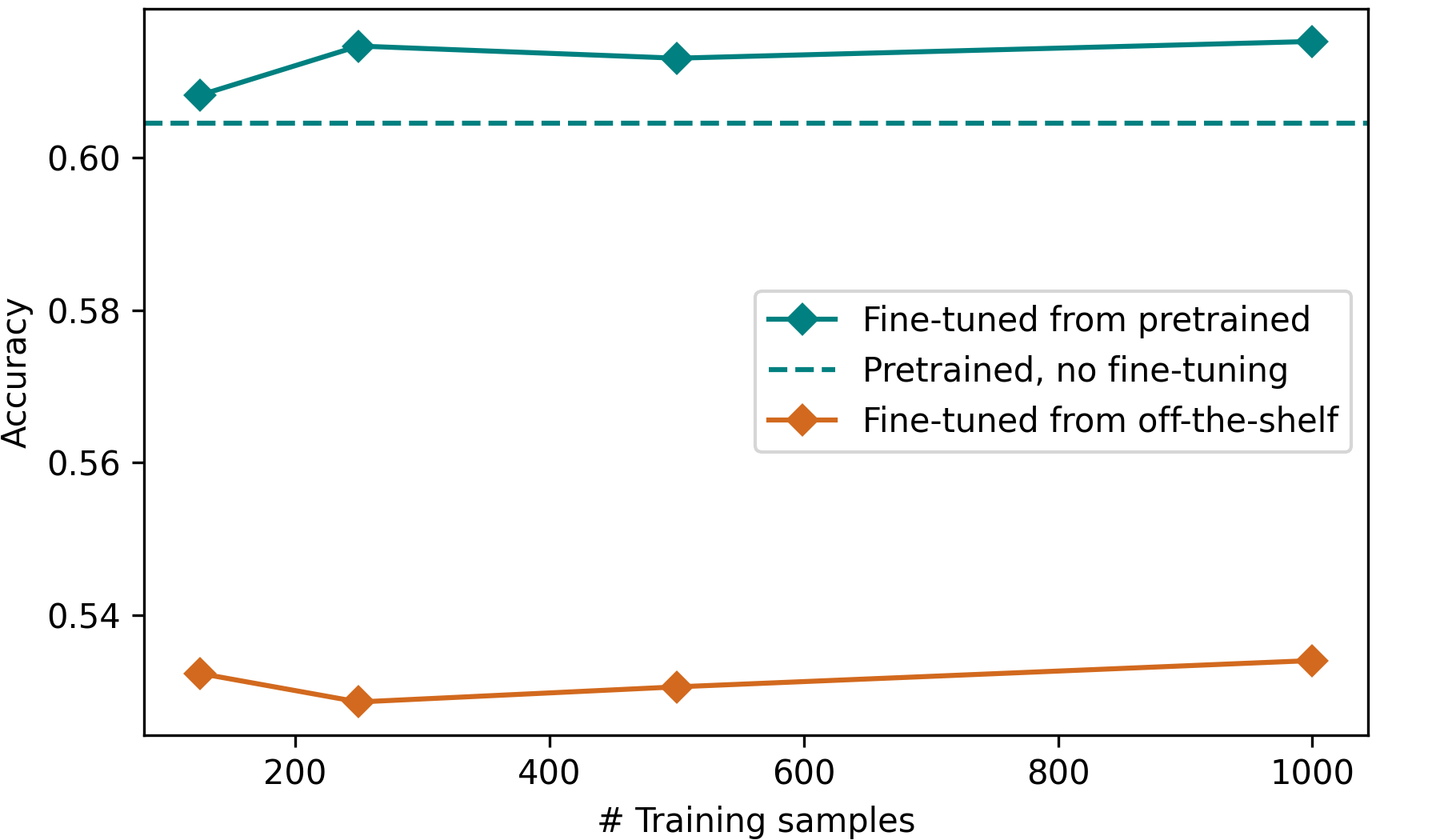}
    \caption{Mean validation accuracy on held-out domains of a RoBERTa+\dsb~model on {\sc arxiv}, fine-tuned using a variable number of random samples from the heldout domain. In our experiments, fine-tuning a contextual embedding model pretrained for the same task on other domains is much better than simply fine-tuning an off-the-shelf model.
    \label{fig:fine-tune-arxiv}}
\end{figure}

Here, we take the best-performing RoBERTa model from section \S\ref{sec:holdout-source}, and fine-tune it with a small number of samples from the unseen domain from the train split in the heldout domain, using a variable number of labeled samples, then evaluate the model using the validation split in the heldout domain. 
Figure \ref{fig:fine-tune-arxiv} demonstrates that even with a relatively small number of labeled samples from the unseen domain, second-pass fine-tuning results in increased performance, but the amount of improvement flattens out as number of samples increases. Of course, users will also need additional data for evaluating in-domain performance, so this underestimates the total amount of labeled data that would be required.

More importantly, we find that fine-tuning a model that has already been trained for the same task on out-of-domain data does far better than fine-tuning a generic off-the-shelf model, even with 1000 in-domain samples. Thus, despite the power of fine-tuning contextual embedding models, there is still a clear advantage for the CSS community of model producers creating such models for measuring categories of interest in text.

\subsection{Comparison to Off-the-shelf Models}
\label{appendix:sentiment}

To ensure that our linear classifiers achieve reasonable performance, we also compare our results on the {\sc senti} dataset to several off-the-shelf sentiment lexicons, evaluating them as classifiers with fine-tuned classification thresholds. As baselines, we evaluate the following off-the-shelf models: VADER \citep{hutto.2014}, LIWC \citep{liwc.2010}, SentiWordNet \citep{baccianella2010sentiwordnet}, a classic Opinion Lexicon \citep{oplex}, and the General Inquirer \citep{stone.1966}.

For each lexicon, we use the available word lists as features, incorporating feature weights when they are provided.
As above, we evaluate all models in comparison to our logistic regression model in terms of out-of-domain performance, working with each domain in the {\sc senti} dataset in turn.

We try using each lexicon both as provided (Untuned), and by introducing a learnable threshold (Tuned). In the latter case, we fine tune the threshold to each target domain in turn, using the same 250 samples from that domain as we use to estimate label distribution for our best model. 

Results are shown in Table \ref{tab:sentiment_results}. Notably, while there is some variation in performance across lexicons (showing the sensitivity of results to which lexicon is chosen), more recent models do not perform markedly better than the General Inquirer from 1962. When fine-tuning to the target domain, none do as well as the logistic model using \dsn~and \dsb, indicating that even commercial lexicons, such as LIWC, are no better at generalizing to new domains than a regularized logistic regression model trained on data from a diverse set of other domains.

\begin{table}[]
    \centering
    \small
    \begin{tabular}{l c  c}
        \toprule    
        Model / Lexicon & Untuned Acc & Tuned Acc \\
        \midrule
        General Inquirer   & 0.635 & 0.675 \\    
        Opinion Lexicon  & \textbf{0.680} & 0.706 \\
        SentiWordNet  & 0.608 & 0.680 \\
        LIWC  & 0.648 & 0.689 \\
        VADER  & 0.631 & - \\
        LogReg  & 0.647 & \textbf{0.712} \\
        \bottomrule
    \end{tabular}
    \caption{Average validation accuracy in unseen domains for several popular off-the-shelf sentiment tools in comparison to our logistic regression model (LogReg). Lexicons are used either as given (Untuned), or with a classification threshold tuned on 250 samples from the target domain (Tuned). For LogReg, Untuned refers to the baseline, and Tuned is the model with \dsn~and \dsb~applied using the same 250 samples to estimate the label distribution. VADER is not tuned as it is distributed as a classifier. 
    }
    \label{tab:sentiment_results}
\end{table}

\section{Discussion and Recommendations} 
\label{sec:discussion}

A key idea of this paper is that domain adaptation should not be something that only model consumers have to confront. Rather, we should think of domain adaptation as a modular, collaborative process, in which model producers should anticipate that model consumers will want to apply models to new domains. 
Ideally, model producers would also make training data available to model consumers, so as to facilitate domain adaptation.
For settings in which this is not possible, we have presented two techniques (\dsb~and \dsn) which improved performance for both logistic regression and contextual embedding models, and we encourage the development of additional techniques.

Although it is still useful for model producers to report performance in the training domain as part of model documentation \citep{mitchell.2019}, model consumers should not rely on such estimates for off-the-shelf models, given the expected performance drop across domains (\citealp{elsahar.2019}; see also Appendix \ref{appendix:perf-drop}). Rather, it is essential to have sufficient labeled data in the application domain to be able to estimate performance, in addition to any labeled data to be used for adaptation, and this should be budgeted for when planning annotations \citep{bai.2021}. For specific applications, model consumers may also care about metrics beyond accuracy, and should evaluate models based on what is most relevant. In addition, these ideas could be fruitfully combined with techniques for lexicon expansion, to account for terms which were not present in the original domain(s) \citep{hamilton.2016,sedinkina.2019}.

Lexicons such as LIWC have an enduring popularity, in part because of their ease of use. As the results above demonstrate, however, simple logistic regression models can do as well (in terms of classification accuracy). Contextual embedding models derived from the same data are considerably more accurate, and need not be any more difficult for practitioners to apply. Thus, we encourage CSS researchers to produce and share such models, even if the raw data itself cannot be shared.

\section{Conclusion}
\label{sec:conclusions}

Using off-the-shelf text classification models for computational social science requires careful thought regarding domain shift. In this paper, we approach this as a modular process in which model producers can apply techniques of \emph{anticipatory domain adaptation} to facilitate adaptation by model consumers. 
We demonstrate that using domain-specific bias (\dsb) and domain-specific normalization (\dsn) produces a reliable performance boost for the model consumers, and that this applies to both linear and contextual embedding models.
Finally, for cases where accuracy is more important than interpretability, we demonstrate the superior out-of-domain performance of contextual embedding models when compared to linear models, even without additional fine-tuning, and encourage model producers to make multiple types of models available.

\section*{Ethical Considerations}

This paper is concerned with possible approaches to domain adaptation, especially for situations where training data cannot be shared, such as for reasons of privacy or copyright. However, it is important to note that domain adaptation will be most effective when model producers are able to make their training data publicly available, and we strongly encourage all researchers to do so, where possible, along with following other best practices for open and reproducible science. 

Although we found significant improvements on out-of-domain data in multiple domains, we only evaluated these techniques on text classification tasks here, and they should therefore be applied with caution. As emphasized throughout the paper, validation is important, especially when using text classification as a form of measurement, and any inferences based on such measurements should be properly contextualized when reporting findings.

Our experiments are all based on pre-established datasets, which do not pose any serious ethical concerns. We also facilitate replication of our results by making code available.

\section*{Acknowledgements}

This research was supported in part by a seed grant from the Stanford Woods Institute for the Environment EVP and by Stanford Data Science. Many thanks to Dan Iter, Mirac Suzgun, Kaitlyn Zhou, and anonymous reviewers for helpful comments and suggestions.

\bibliography{acl2020}
\bibliographystyle{acl_natbib}

\clearpage

\onecolumn
\appendix

\section{Full Heldout Domain Accuracy}
\label{appendix:full-holdout-source}

For each model-technique combination, for each dataset, and for each domain in the dataset, we train a model using the training split of all domains except the single heldout domain, then evaluate the model on the heldout domain, then average accuracy across these domains. These data were used to determine which model comparisons to test for significance, though we include all results on test data in the main paper for completeness.

\begin{table}[H]
\centering
\small
\begin{tabular}{llllllllll}
\toprule
&& \multicolumn{2}{c}{\sc mfc} & \multicolumn{2}{c}{\sc arxiv} 
& \multicolumn{2}{c}{\sc amazon} & \multicolumn{2}{c}{\sc senti} \\
&& \multicolumn{1}{c}{acc} & \multicolumn{1}{c}{$\sigma_\Delta$} 
& \multicolumn{1}{c}{acc} & \multicolumn{1}{c}{$\sigma_\Delta$}
& \multicolumn{1}{c}{acc} & \multicolumn{1}{c}{$\sigma_\Delta$}
& \multicolumn{1}{c}{acc} & \multicolumn{1}{c}{$\sigma_\Delta$}\\ \midrule
\multirow{8}{*}{LogReg} 
 & Base     & 0.501 & -     & 0.541 & -     & 0.672 & -     & 0.647 &  - \\ 
  & DR       & 0.493 & 0.006 & 0.552 & 0.005 & 0.674 & 0.004 & 0.648 & 0.003 \\
  & GR       & 0.502 & 0.002 & 0.542 & 0.003 & 0.709 & 0.001 & 0.638 & 0.003 \\
  & \dsn      & 0.452 & 0.013 & 0.483 & 0.033 & 0.682 & 0.012 & 0.595 & 0.044  \\
  & \dsb~(oracle)   & 0.520 & 0.020  & 0.565 & 0.014 & 0.715 & 0.003 & 0.695 & 0.041 \\
 & \dsb+\dsn~(oracle) & 0.536 & 0.017 & 0.570 & 0.013 & 0.717 & 0.002 & 0.712 & 0.039 \\
\midrule
\multirow{4}{*}{RoBERTa} 
 & Base     & 0.581 & -     & 0.583 & -     & 0.772 &  -    & 0.803 & - \\
 & DR       & 0.585 & 0.014 & 0.587 & 0.005 & 0.782 & 0.017 & 0.817 & 0.012 \\
 & GR       & 0.204 & 0.046 & 0.510 & 0.010  & 0.778 & 0.012 & 0.684 & 0.068 \\
 & \dsb~(oracle)     & 0.615 & 0.031 & 0.605 & 0.011 & 0.779 & 0.012 & 0.819 & 0.014 \\
\bottomrule
\end{tabular}
\caption{Out-of-domain accuracy of models trained holding out one domain per trial, then evaluated on the heldout domain, for all configurations of each model. $\sigma_\Delta$ is the standard deviation of accuracy difference in each domain over the corresponding baseline (``Base'').}
\label{tab:full-holdout-source}
\end{table}

\section{Single Domain Training}
\label{appendix:extreme-data-scarcity}

Similar to the previous experiment where we held out a single domain, here we train only on a single domain, and evaluate with all non-training domains.

\begin{table}[H]
\centering
\small
\begin{tabular}{llllllllll}
\toprule
&& \multicolumn{2}{c}{\sc mfc} & \multicolumn{2}{c}{\sc arxiv} 
& \multicolumn{2}{c}{\sc amazon} & \multicolumn{2}{c}{\sc senti} \\
&& \multicolumn{1}{c}{acc} & \multicolumn{1}{c}{$\sigma_\Delta$} 
& \multicolumn{1}{c}{acc} & \multicolumn{1}{c}{$\sigma_\Delta$}
& \multicolumn{1}{c}{acc} & \multicolumn{1}{c}{$\sigma_\Delta$}
& \multicolumn{1}{c}{acc} & \multicolumn{1}{c}{$\sigma_\Delta$}\\ \midrule
\multirow{8}{*}{LogReg} 
 & Base     & 0.426 & -      & 0.555 & -      & 0.653 & -     & 0.574 & -  \\
  & DR
 & 0.423 & 0.002  & 0.574 & 0.012  & 0.605 & 0.002 & 0.571 & 0.006  \\ 
 & GR
 & 0.425 & 0.000    & 0.554 & 0.000    & 0.652 & 0.001 & 0.572 & 0.002  \\
  & \dsn      & 0.366 & 0.010   & 0.417 & 0.019  & 0.629 & 0.015 & 0.545 & 0.013  \\
 & \dsb~(oracle)      & 0.447 & 0.006  & 0.596 & 0.008  & 0.681 & 0.016 & 0.670 & 0.018  \\
 & \dsb+\dsn~(oracle)  & 0.472 & 0.008  & 0.598 & 0.007  & 0.683 & 0.015 & 0.670 & 0.018  \\
\midrule
\multirow{4}{*}{RoBERTa}
 & Base     & 0.48  & -      & 0.539 & -      & 0.727 & -     & 0.622 & -  \\
 & DR
 & 0.510 & 0.023  & 0.542 & 0.004  & 0.736 & 0.028 & 0.620 & 0.014  \\
 & GR
 & 0.168 & 0.034  & 0.448 & 0.074  & 0.647 & 0.026 & 0.548 & 0.062  \\
 & \dsb~(oracle)    & 0.540 & 0.029  & 0.560 & 0.008  & 0.751 & 0.023 & 0.699 & 0.039  \\
\bottomrule
\end{tabular}
\caption{Out-of-domain accuracy of models trained with a single domain, then evaluated on all other domains combined, for all configurations of each model. $\sigma_\Delta$ is the standard deviation of accuracy difference in each domain over the corresponding baseline (Base).}
\label{tab:extreme-data-scarcity}
\end{table}

In single domain training, since no deconfounding between training domain is possible, gradient reversal (GR) and deep residualization (DR) fails to meaningfully improve performance.
 
Comparing table \ref{tab:extreme-data-scarcity} to table \ref{tab:full-holdout-source}, not only do we observe a very similar trend of performance differences, where our recommended model-technique combinations (LogReg+\dsb+\dsn, RoBERTa+\dsb) consistently outperforms the rest, but the difference is more pronounced. 

\section{Out-of-domain Performance Drop}
\label{appendix:perf-drop}

\begin{table}[H]
\centering
\small
\begin{tabular}{@{}ccccccccccccc@{}}
\toprule
 & \multicolumn{3}{c}{\sc MFC} & \multicolumn{3}{c}{\sc arXiv} 
 & \multicolumn{3}{c}{\sc amazon} & \multicolumn{3}{c}{\sc senti}\\
 & ID & OOD & $\sigma_\Delta$ & ID & OOD & $\sigma_\Delta$
 & ID & OOD & $\sigma_\Delta$ & ID & OOD & $\sigma_\Delta$\\ \midrule
LogReg  & 0.607 & 0.508 & 0.036 & 0.583 & 0.542 & 0.012 & 0.722 & 0.672 & 0.062 & 0.756 & 0.649 & 0.060 \\
RoBERTa & 0.703 & 0.600 & 0.071 & 0.608 & 0.571 & 0.021 & 0.797 & 0.772 & 0.021 & 0.837 & 0.789 & 0.073\\ \bottomrule
\end{tabular}
\caption{Test accuracy of models trained on all domains then evaluated on the test split of each domain (in-domain ``ID"), and trained on all but one held-out domain then evaluated on the test split of that held-out domain (out-of-domain ``OOD''). $\sigma_\Delta$ is the standard deviation of accuracy difference across domains. }
\label{tab:perf-drop}
\end{table}

\clearpage

\section{Estimating Performance}
\label{appendix:estimate-performance}

\begin{figure}[H]
    \centering
    \includegraphics[width=1.0\textwidth]{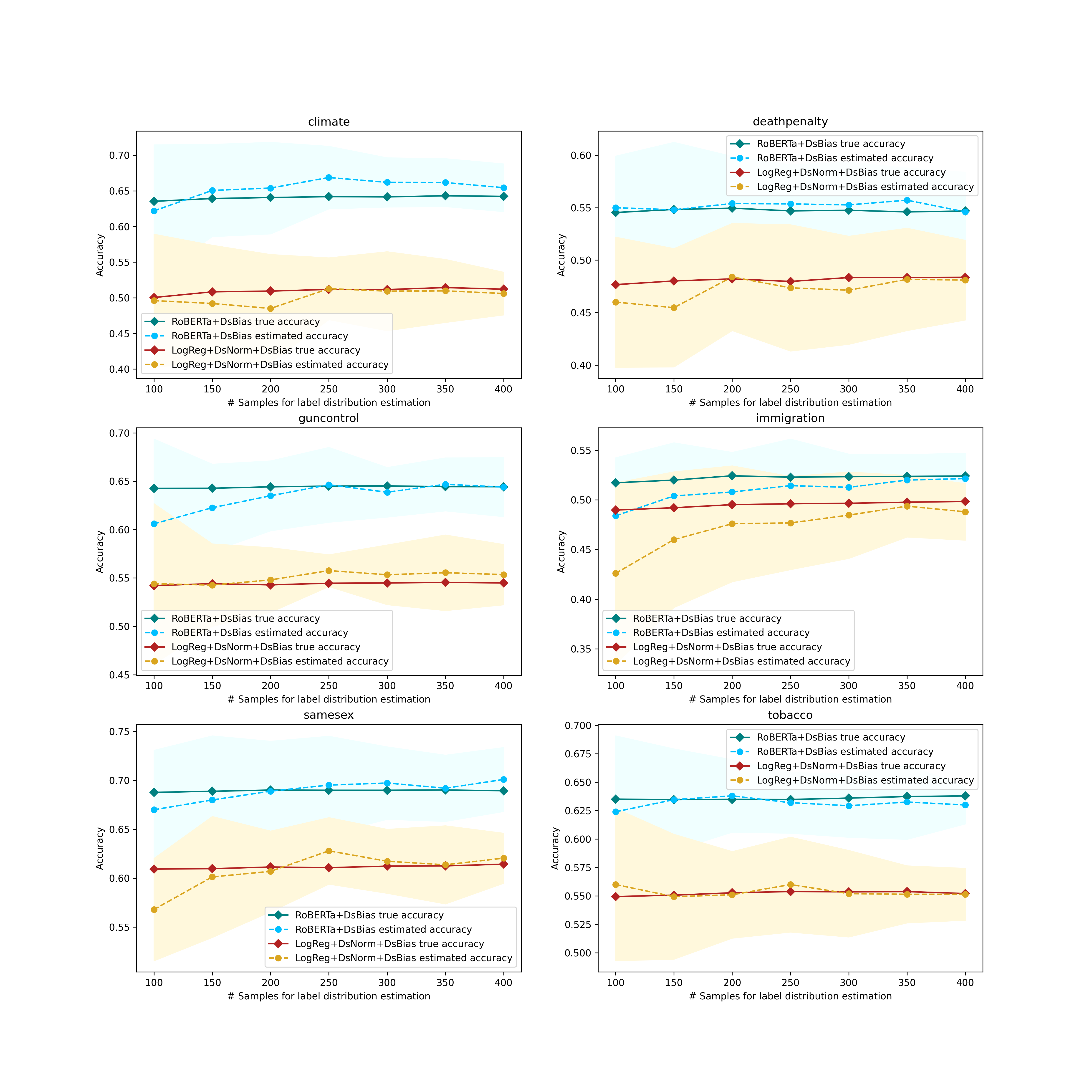}
    \caption{Validation accuracy calculated from all holdout samples, and from limited samples, of each topic (domain) in the Media Frame Corpus ({\sc mfc}). Shaded area denotes 1 standard deviation from mean estimated performance}
    \label{fig:performance-est-full-frame}
\end{figure}

\clearpage

\begin{figure}[H]
    \centering
    \includegraphics[width=1.0\textwidth]{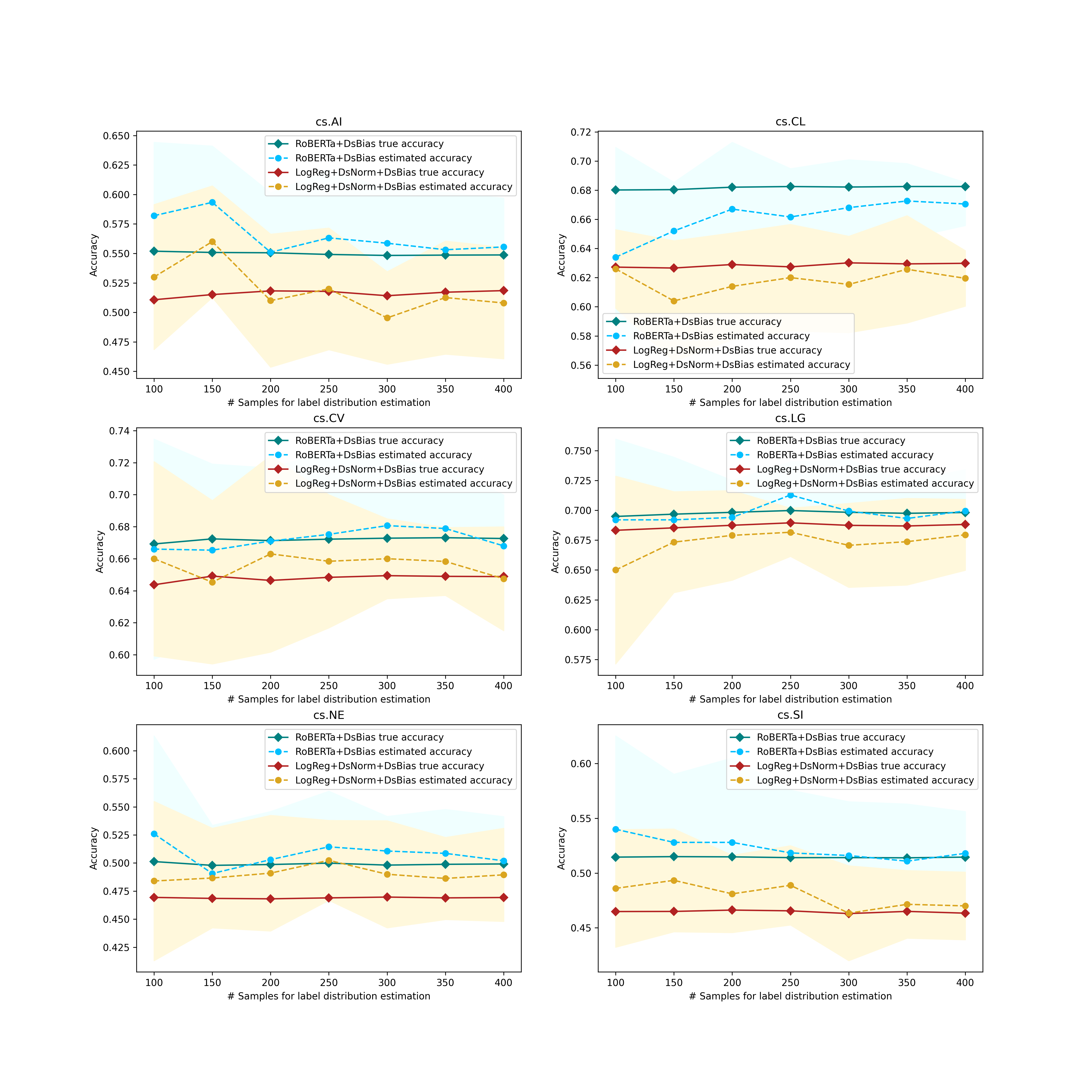}
    \caption{Validation accuracy calculated from all holdout samples, and from limited samples, of each category (domain) in {\sc arxiv}. Shaded area denotes 1 standard deviation from mean estimated performance} 
    \label{fig:performance-est-full-arxiv}
\end{figure}

\clearpage

\begin{figure}[H]
    \centering
    \includegraphics[width=1.0\textwidth]{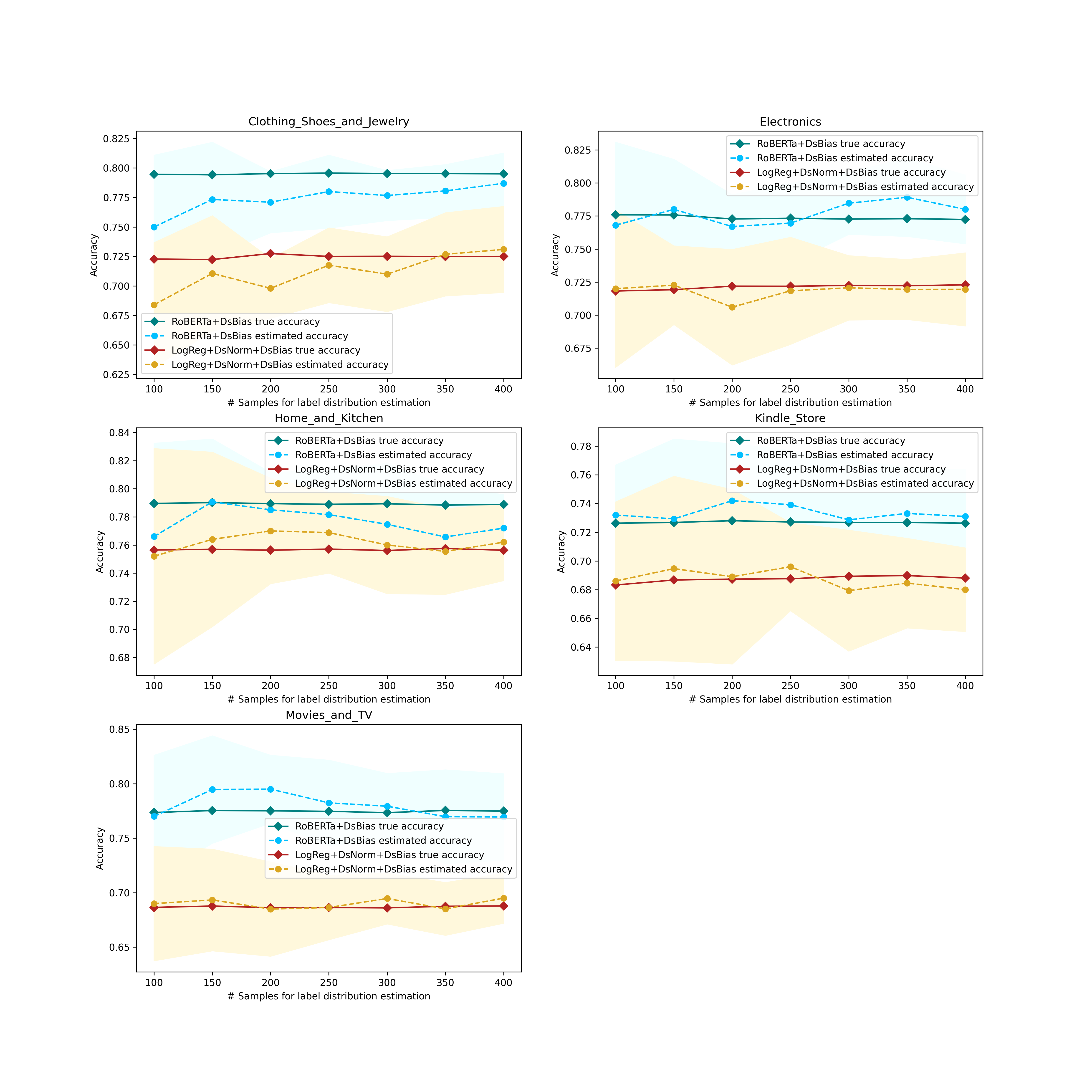}
    \caption{Validation accuracy calculated from all holdout samples, and from limited samples, of each category (domain) in {\sc amazon}. Shaded area denotes 1 standard deviation from mean estimated performance} 
    \label{fig:performance-est-full-amazon}
\end{figure}

\clearpage

\begin{figure}[H]
    \centering
    \includegraphics[width=1.0\textwidth]{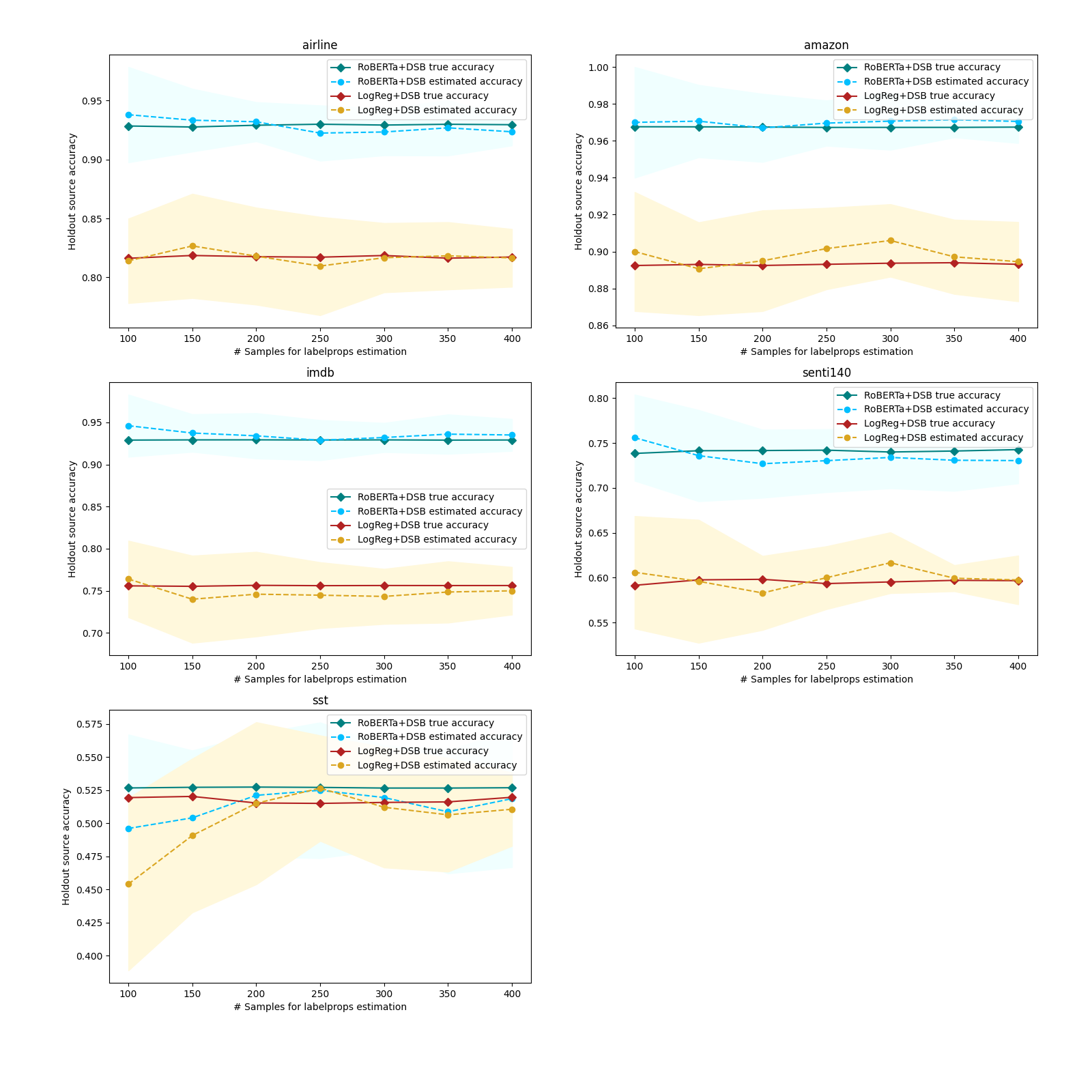}
    \caption{Validation accuracy calculated from all holdout samples, and from limited samples, of each sub-dataset (domain) in {\sc senti}. Shaded area denotes 1 standard deviation from mean estimated performance} 
    \label{fig:performance-est-full-senti}
\end{figure}

\clearpage

\section{Example Lexicon}

\label{appendix:lexicon}

\begin{table}[H]
\centering
\small
\begin{tabular}{@{}cccccccc@{}}
\toprule
Economic & \begin{tabular}[c]{@{}c@{}}Capacity \\ and \\ Resources\end{tabular} & Morality & \begin{tabular}[c]{@{}c@{}}Fairness \\ and \\ Equality\end{tabular} & \begin{tabular}[c]{@{}c@{}}Legality, \\ Constitutionality, \\ Jurisdiction\end{tabular} & \begin{tabular}[c]{@{}c@{}}Policy \\ Prescription \\ and \\ Evaluation\end{tabular} & \begin{tabular}[c]{@{}c@{}}Crime \\ and \\ Punishment\end{tabular} & \begin{tabular}[c]{@{}c@{}}Security \\ and \\ Defense\end{tabular} \\ \midrule
economic & applications & moral & discrimination & asylum & ordinance & criminals & terrorist \\
financial & shortage & church & fairness & lawsuit & rid & deport & security \\
budget & species & pope & black & justices & punishment & deported & terrorists \\
business & capacity & catholic & equality & sued & vehicles & allegedly & border \\
economy & ocean & churches & innocent & suing & policy & injection & military \\
fund & handle & leaders & race & constitution & penalty & minors & patrol \\
jobs & process & christian & racial & plaintiffs & citizenship & smuggling & fbi \\
costs & surge & religious & equal & lawsuits & effect & kill & terror \\
economists & science & rev & innocence & visa & plan & crackdown & threats \\
sales & resources & francis & evidence & suit & bill & deportation & pentagon \\
corporate & scientists & bishop & unfair & court & ban & fine & intelligence \\
company & foreign & faith & fair & visas & would & police & terrorism \\
companies & wait & rabbi & blacks & judge & policies & investigators & protect \\
tax & critical & churchs & testimony & attorney & smokefree & firstdegree & guard \\
cost & waiting & jewish & facts & antonin & proposal & prison & war \\
revenue & years & society & civil & militia & bans & maximum & secure \\
stores & tons & clergy & racist & shall & supporters & arrested & airports \\
treasury & growing & christians & true & lawyers & designated & sentenced & attacks \\
dollars & used & nicotine & equally & licenses & buildings & scheme & russian \\
money & lines & bible & treated & granted & homeland & executed & defense \\ \bottomrule
\end{tabular}

\vspace{2em}

\begin{tabular}{@{}ccccccc@{}}
\toprule
\begin{tabular}[c]{@{}c@{}}Health \\ and \\ Safety\end{tabular} & \begin{tabular}[c]{@{}c@{}}Quality \\ of \\ Life\end{tabular} & \begin{tabular}[c]{@{}c@{}}Cultural \\ Identity\end{tabular} & \begin{tabular}[c]{@{}c@{}}Public \\ Sentiment\end{tabular} & Political & \begin{tabular}[c]{@{}c@{}}External \\ Regulation\\ and \\ Reputation\end{tabular} & Other \\ \midrule
mentally & daughter & documentary & poll & governor & countries & hillary \\
health & loved & film & protesters & republicans & minister & chris \\
condition & benefits & movie & rally & bloombergs & mexican & gop \\
medical & quit & culture & protest & conservatives & foreign & annual \\
disease & mother & actor & marched & sen & european & paid \\
doctors & weather & cultural & demonstrators & clinton & un & brother \\
suicide & college & book & voters & reelection & mexicans & cultural \\
hospital & families & ethnic & activists & bipartisan & visit & money \\
pain & tears & executions & organizers & gop & france & supporting \\
safe & temperatures & population & organized & mayor & states & stores \\
safety & felt & english & gathered & hillary & china & accused \\
mental & family & movies & protests & statements & negotiations & interests \\
lung & everything & history & mom & rep & agreement & governors \\
coverage & temperature & players & polls & cuomo & united & candidate \\
locks & living & tv & polling & mayors & talks & fund \\
retarded & married & census & mothers & endorsement & mexico & endorsement \\
lungs & conditions & league & attitudes & obama & summit & didnt \\
risk & life & decline & nra & referendum & australia & economic \\
illness & classes & star & signatures & ryan & mexicos & reelection \\
diseases & father & smoked & organization & republican & canadian & shortly \\ \bottomrule
\end{tabular}

\caption{Top weighted 20 words from each class in a lexicon elicited from the Media Frame Corpus ({\sc mfc}), with a logistic regression model and using Domain-Specific Bias (\dsb) and Domain-Specific Normalization (\dsn). Weight value associated with each word not included. }
\label{tab:example-lexicon-mfc}
\end{table}

\clearpage

\begin{table}[H]
\centering
\small

\begin{tabular}{@{}cccc@{}}
\toprule
-2008 & 2009-2014 & 2015-2018 & 2019- \\ \midrule
rules & web & recurrent & covid19 \\
grammar & bayesian & deep & bert \\
presented & belief & convolutional & federated \\
logic & variables & neural & transformer \\
described & markov & lstm & selfsupervised \\
grammars & graphical & big & fewshot \\
theory & svm & adversarial & pandemic \\
statistical & technique & pascal & transformerbased \\
describes & probabilistic & endtoend & fairness \\
parsing & words & embeddings & selfattention \\
information & propagation & reinforcement & sota \\
linguistic & probabilities & nonconvex & transformers \\
general & convex & stateoftheart & ai \\
syntactic & recognition & dataset & explainable \\
disambiguation & svms & propose & downstream \\
shown & database & sentiment & explainability \\
sense & independence & convnet & outofdistribution \\
definition & conditional & stochastic & nas \\
discussed & uncertainty & mnist & learningbased \\
tested & basis & dropout & embeddings \\
class & immune & atari & code \\
notion & em & rnn & backbone \\
semantics & sparse & sequencetosequence & gnns \\
presents & dictionary & generative & gnn \\
programming & wavelet & train & augmentation \\
programs & sound & gradient & quantum \\
order & collaborative & embedding & continual \\
algorithm & extraction & convnets & lightweight \\
classes & management & explore & neural \\
two & coding & machine & unet \\
noun & techniques & jointly & module \\ \bottomrule
\end{tabular}

\caption{Top weighted 30 words from each class in a lexicon elicited from the abstract texts in the arXiv dataset ({\sc arXiv}), with a logistic regression model and using Domain-Specific Bias (\dsb) and Domain-Specific Normalization (\dsn). Weight value associated with each word not included. }
\label{tab:example-lexicon-arxiv}
\end{table}

\clearpage

\begin{table}[H]
\centering
\small

\begin{tabular}{@{}ccc@{}}
\toprule
Negative (1 star) & Neutral (2-4 stars) & Positive (5 stars) \\ \midrule
waste & ok & love \\\
poor & stars & perfect \\\
junk & okay & excellent \\\
horrible & however & awesome \\\
terrible & disappointing & loves \\\
worst & otherwise & perfectly \\\
awful & unfortunately & great \\\
return & complaint & highly \\\
returned & overall & glad \\\
cheaply & downside & loved \\\
useless & returned & amazing \\\
boring & bit & pleased \\\
poorly & reason & beautiful \\\
broke & cute & thank \\\
garbage & returning & wonderful \\\
disappointed & little & thanks \\\
nothing & wish & happy \\\
disappointing & though & fantastic \\\
died & good & favorite \\\
apart & slow & comfortable \\\
cheap & decent & compliments \\\
crap & flimsy & wait \\\
defective & annoying & gorgeous \\\
refund & stiff & exactly \\\
returning & runs & best \\\
money & issue & worried \\\
month & liked & admit \\\
beware & missing & happier \\\
uncomfortable & interesting & wow \\\
fell & nice & worry \\\
stopped & alright & adorable \\\
star & overpriced & faster \\\
disappointment & except & nice \\\
completely & problem & helps \\\
weak & expected & incredible \\\
description & awkward & classic \\\
even & gave & satisfied \\\
bad & thinner & originally \\\
within & flaw & charm \\\
minutes & cons & classy \\\
broken & concept & durable \\\
cannot & sometimes & needed \\\
shame & seems & fast \\\
worse & mechanism & comfy \\\
unless & bulky & beautifully \\\
piece & lack & truly \\\
barely & pretty & recently \\\
stuck & narrow & easier \\\
ripped & meh & ram \\\
please & careful & cleans \\\
\\ \bottomrule
\end{tabular}

\caption{Top weighted 50 words from each class in a lexicon elicited from amazon review texts ({\sc amazon}), with a logistic regression model and using Domain-Specific Bias (\dsb) and Domain-Specific Normalization (\dsn). Weight value associated with each word not included. }
\label{tab:example-lexicon-amazon}
\end{table}

\clearpage

\begin{table}[H]
\centering
\small

\begin{tabular}{@{}cc@{}}
\toprule
Negative & Positive \\ \midrule
poorly & thank \\
annoying & thanks \\
worst & superb \\
boring & hi \\
hurts & amazing \\
waste & brilliant \\
dislike & excellent \\
ugh & subtle \\
finale & smooth \\
disappointed & awesome \\
sad & wonderfully \\
poor & outstanding \\
wooden & hahaha \\
redeeming & yay \\
cancelled & excited \\
sucks & hilarious \\
wanna & notice \\
disappointment & seemingly \\
bag & funniest \\
unfortunately & safe \\
ugly & noir \\
mediocre & impressed \\
laughable & extraordinary \\
crappy & haha \\
lousy & powerful \\
turkey & humorous \\
claims & loved \\
sorry & solid \\
junk & helpful \\
arms & higher \\
sick & germany \\
awful & dvd \\
disappointing & ideal \\
pointless & sweet \\
shots & twenty \\
barely & great \\
confused & pleasure \\
headache & friday \\
ruined & happy \\
ticket & independent \\
potential & involve \\
obnoxious & masterpiece \\
luggage & captures \\
shallow & welcome \\
pain & rare \\
anymore & cool \\
nowhere & south \\
terrible & incredible \\
miss & best \\
min & gripping \\
\\ \bottomrule
\end{tabular}

\caption{Top weighted 50 words from each class in a lexicon elicited from a collection of multiple sentiment classification datasets ({\sc senti}), with a logistic regression model and using Domain-Specific Bias (\dsb) and Domain-Specific Normalization (\dsn). Weight value associated with each word not included. }
\label{tab:example-lexicon-senti}
\end{table}

\clearpage

\section{Data Splits}

\label{appendix:data-split}

For the Media Frame Corpus {\sc (mfc)}, we a fixed number of 400 random samples from each news issue (domain) as the test set, and do not use them for any training or hyperparameter tuning until the end for reporting test performance. Validation data for hyperparameter tuning in experiments is either from a held-out source, or k-fold validation.

\begin{table}[H]
\centering
\small
\begin{tabular}{@{}cccccccc@{}}
\toprule
 & Climate & Gun control & Death penalty & Immigration & \begin{tabular}[c]{@{}c@{}}Same-sex \\ marriage\end{tabular} & Tobacco & Total \\ \midrule
Train & 3795 & 3777 & 8498 & 5533 & 3956 & 3251 & 28810 \\
Test & 400 & 400 & 400 & 400 & 400 & 400 & 2400 \\
Total & 4195 & 4177 & 8898 & 5933 & 4356 & 3651 & 31210 \\ \bottomrule
\end{tabular}
\caption{Sample sizes of each domain and each split from the Media Frame Corpus {(\sc mfc)}}
\label{tab:sample-size-mfc}
\end{table}

For the arXiv dataset {\sc (arXiv)}, we take a fixed proportion of 10\% of random samples from each paper category (domain) as the test set, and do not use them for any training or hyperparameter tuning until the end for reporting test performance. Validation data for hyperparameter tuning in experiments is either from a held-out source, or k-fold validation. 

\begin{table}[H]
\centering
\small
\begin{tabular}{@{}cccccccc@{}}
\toprule
 & \begin{tabular}[c]{@{}c@{}}Artificial\\ intelligence\\ (cs.AI)\end{tabular} & \begin{tabular}[c]{@{}c@{}}Computation\\ and\\ language\\ (cs.CL)\end{tabular} & \begin{tabular}[c]{@{}c@{}}Computer\\ vision\\ (cs.CV)\end{tabular} & \begin{tabular}[c]{@{}c@{}}Machine\\ learning\\ (cs.LG)\end{tabular} & \begin{tabular}[c]{@{}c@{}}Neural \\ and \\ evolutionary \\ computing\\ (cs.NE)\end{tabular} & \begin{tabular}[c]{@{}c@{}}Social \\ and \\ Information \\ Networks\\ (cs.SI)\end{tabular} & Total \\ \midrule
Train & 18294 & 21131 & 46008 & 53647 & 4798 & 11086 & 154986 \\
Test & 2034 & 2350 & 5113 & 5962 & 534 & 1233 & 17226 \\
Total & 20328 & 23481 & 51121 & 59609 & 5332 & 12319 & 172212 \\ \bottomrule
\end{tabular}
\caption{Sample sizes of each domain and each split from the arXiv dataset {\sc (arXiv)}}
\label{tab:sample-size-arxiv}
\end{table}

For the Amazon reviews dataset {\sc amazon}, we first subsample to keep only 0.2\% of the original dataset size to simulate a data-scarce setting. We then take a fixed proportion of 10\% of random samples from each category (domain) as the test set, and do not use them for any training or hyperparameter tuning until the end for reporting test performance. Validation data for hyperparameter tuning in experiments is either from a held-out source, or k-fold validation.

\begin{table}[H]
\centering
\small
\begin{tabular}{@{}ccccccc@{}}
\toprule
 & Clothing, Shoes and Jewelry
 & Electronics
 & Home and Kitchen
 & Kindle Store
 & Movies and TV
 & Total\\ \midrule
Train & 20315 & 12132 & 	12418 & 	4002 & 	6140 & 	55007 \\
Test & 2258 & 1350 & 	1382 & 	446 & 	683 & 	6119 \\
Total & 22573 & 13482 & 	13800 & 	4448 & 	6823 & 	61126 \\\bottomrule
\end{tabular}
\caption{Sample sizes of each domain and each split from the Amazon review dataset {\sc (amazon)}}
\label{tab:sample-size-amazon}
\end{table}

For {\sc senti}, we take a fixed proportion of 10\% of random samples from each data source (domain) as the test set, and do not use them for any training or hyperparameter tuning until the end for reporting test performance. Validation data for hyperparameter tuning in experiments is either from a held-out source, or k-fold validation.

\begin{table}[H]
\centering
\small
\begin{tabular}{@{}ccccccc@{}}
\toprule
 & Airline Tweets
 & Amazon Books
 & IMDb Movie Reviews
 & Sentiment 140
 & \begin{tabular}[c]{@{}c@{}}Stanford Sentiment \\ Treebank \end{tabular}
 & Total\\ \midrule
Train & 7080& 	7843& 	8977& 	9002& 	2778 & 35680 \\
Test & 788& 	873& 	999& 	1001& 	310 & 3971 \\
Total & 7868& 	8716& 	9976& 	10003& 	3088 & 39651\\
\bottomrule
\end{tabular}
\caption{Sample sizes of each domain and each split from the sentiment classification dataset collection {\sc (senti)}}
\label{tab:sample-size-senti}
\end{table}

\section{Data Preprocessing}

Sample texts are preprocessed before used to train models and perform experiments. For both types of models, urls are first removed from the text. If the text is from a Tweet, then Twitter handles (tokens starting with $@$) and emojis are also identified and removed. 

For RoBERTa models, this sanitized text is then passed into a tokenized as-is without any additional processing. For logistic regression models, we then build a bag-of-word feature vector by first removing all punctuation, special symbols, English stopwords (from NLTK), pure numbers, and tokens including both alphabetical and numeric characters. Finally, we build a vocabulary of a fixed size of 5000 most frequent tokens, and convert the preprocessed texts into feature vectors.

\section{Experiment Setup and Hyperparameter Tuning}
\label{appendix:hyperparams}

As in section \S\ref{sec:holdout-source} and section \S\ref{sec:adapt} we train multiple models of various configurations using different combination of training domains, we maintain a consistent strategy for hyperparameter tuning to ensure performance comparability. 

\paragraph{Logistic regression models} have one hyperparameter, the L1 regularization constant $\lambda$. For each experiment and each model configuration, we first run k-fold validation within the train set, and conduct a search for $\lambda = 1^{-5} \times 2^k, k\in (0,4)$, while optimizing for lowest loss on the main prediction target on the validation set. Then we use the same optimal $\lambda$ to train with the full train set until convergence. 

\paragraph{RoBERTa models} have one hyperparameter, the number of epochs $E$ to train or fine-tune. Since deep contextual embedding models are very powerful in the context of our small datasets, we early-stop during training to ensure it does not overfit to the training data. For each experiment and each model configuration, we first run k-fold validation within the train set, and conduct a search for $E \in (1,8)$ for the out-of-domain experiments, and for $E \in (1,15)$ the domain fine-tuning experiments, while optimizing for lowest loss on the main prediction target on the validation set. Then we use the full train set and train for the same optimal $E$ epochs.

\section{Power Analysis}
\label{appendix:power-analysis}

Prior to testing for significant differences between models, as reported in the main paper (\S\ref{sec:holdout-source}), we conduct a simple power analysis using the results obtained on validation data (Appendix \ref{appendix:full-holdout-source}), to ensure that such tests will be adequately powered. To do so, we follow the approach described in \citet{card.2020}, basing our calculation on the estimated differences in accuracy and rates of agreement between pairs of models on validation data.

Results are given in Table \ref{tab:power-analysis}. All comparisons are well powered for the improvement of \dsb~on RoBERTa models, and all differences (on test data) are significant. The same is true for comparing the combined effect of \dsb+\dsn~on LogReg models, except on the {\sc amazon}
dataset, but most comparisons for the  improvement from \dsn~alone would be underpowered. 

\begin{table}[H]
\centering
\small

\begin{tabular}{lcccccc}
\toprule
Model A 
& \multicolumn{2}{l}{LogReg}         
& \multicolumn{2}{l}{LogReg+\dsb}     
& \multicolumn{2}{l}{RoBERTa}     \\
Model B 
& \multicolumn{2}{l}{LogReg+\dsb+\dsn} 
& \multicolumn{2}{l}{LogReg+\dsb+\dsn} 
& \multicolumn{2}{l}{RoBERTa+\dsb} \\ \midrule
&  Power & McNemar’s $p$
&  Power & McNemar’s $p$ 
&  Power & McNemar’s $p$ \\ \midrule
{\sc mfc}    & 1.00  & $<0.001$   &  0.36   &  --   & 0.91 & ~$\phantom{<}0.009$   \\
{\sc arXiv}  & 1.00 & $<0.001$    & 0.28 & --  & 1.00 & $<0.001$   \\
{\sc amazon} & 0.49 & --   &  0.41 & --   & 0.95  & $<0.001$ \\
{\sc senti}   & 1.00  & $<0.001$   & 0.97  & $<0.001$ &  0.93 & $<0.001$   \\ \bottomrule
\end{tabular}

\caption{Power analysis results for evaluating potential model comparisons. Statistical power is calculated per \citet{card.2020} using all out-of-domain validation samples, with dataset size equivalent to that of the test split. McNemar's $p$ is reported here using the out-of-domain test data (to evaluate if the difference is significant) for those comparisons that are well powered.  }
\label{tab:power-analysis}
\end{table}

\end{document}